\documentclass{article}

\usepackage{arxiv}

\usepackage[utf8]{inputenc} % allow utf-8 input
\usepackage[T1]{fontenc}    % use 8-bit T1 fonts
\usepackage{hyperref}       % hyperlinks
\usepackage{url}            % simple URL typesetting
\usepackage{booktabs}       % professional-quality tables
\usepackage{amsfonts}       % blackboard math symbols
\usepackage{nicefrac}       % compact symbols for 1/2, etc.
\usepackage{microtype}      % microtypography
\usepackage{lipsum}

\title{Applications of Artificial Intelligence to aid detection of dementia: a narrative review on current capabilities and future directions}
\author{
  Renjie Li \\
  Discipline of Information and Communication Technology \\
  University of Tasmania \\
  TAS 7005 Australia \\
  \texttt{renjie.li@utas.edu.au} \\
  \And
  Xinyi Wang \\
  Discipline of Information and Communication Technology \\
  University of Tasmania \\
  TAS 7005 Australia \\
  \texttt{xinyi.wang@utas.edu.au} \\
  \AND
  Katherine Lawler \\
  Wicking Dementia Research and Education Centre \\
  University of Tasmania \\
  TAS 7000 Australia \\
  \texttt{katherine.lawler@utas.edu.au} \\
  \AND
  Saurabh Garg \\
  Discipline of Information and Communication Technology \\
  University of Tasmania \\
  TAS 7005 Australia \\
  \texttt{saurabh.garg@utas.edu.au} \\
  \AND
  Quan Bai \\
  Discipline of Information and Communication Technology \\
  University of Tasmania \\
  TAS 7005 Australia \\
  \texttt{quan.bai@utas.edu.au} \\
  \AND
  Jane Alty \\
  Wicking Dementia Research and Education Centre \\
  University of Tasmania \\
  TAS 7000 Australia \\
  \texttt{jane.alty@utas.edu.au} \\
}

\begin{document}
\maketitle

\begin{abstract}
With populations ageing, the number of people with dementia worldwide is expected to triple to 152 million by 2050. Seventy percent of cases are due to Alzheimer's disease (AD) pathology and there is a 10-20 year 'pre-clinical' period before significant cognitive decline occurs. We urgently need, cost effective, objective methods to detect AD, and other dementias, at an early stage. Risk factor modification could prevent 40\% of cases and drug trials would have greater chances of success if participants are recruited at an earlier stage. 

Currently, detection of dementia is largely by pen and paper cognitive tests but these are time consuming and insensitive to pre-clinical phases. Specialist brain scans and body fluid biomarkers can detect the earliest stages of dementia but are too invasive or expensive for widespread use. With the advancement of technology, Artificial Intelligence (AI) shows promising results in assisting with detection of early-stage dementia. Existing AI-aided methods and potential future research directions are reviewed and discussed. 
\end{abstract}


\section{Introduction}
Dementia is “the greatest global challenge for health and social care in the 21st century”~\cite{livingston2017dementia}. It describes a degenerative pathological condition of the brain that results in a syndrome of impaired memory, thinking, behaviour and function~\cite{world2019risk}. Dementia is associated with progressive difficulties performing daily activities and reduced quality of life~\cite{world2019risk}. More than 50 million people around the world live with dementia and this figure is expected to rise to 152 million by 2050, driven primarily by ageing populations~\cite{livingston2020dementia}. Dementia devastates families, is a huge burden on healthcare systems, and costs more than US \$1 trillion annually~\cite{livingston2020dementia}. 

Alzheimer's disease (AD) is the most common cause of dementia and the underlying pathology progresses silently in the brain for 10-20 years before cognitive symptoms occur. This pre-clinical stage, and the subsequent mild cognitive impairment (MCI) stage of AD are critical to identify because pathology in the brain is minimal and therefore neuroprotective interventions, such as drug trials and risk reduction, have the greatest chance of success. Early modification of lifestyle (e.g., obesity) and medical (e.g., hypertension) risk factors could prevent 40\% of dementia cases~\cite{livingston2020dementia}. 
Currently, we lack accessible population-level tests to detect pre-clinical AD, MCI or the earliest stages of AD - before significant cognitive and functional decline occur. Currently, AD is usually diagnosed when cognitive symptoms such as memory impairment appear, after more than 20 years of progressive brain pathology. Symptoms of AD gradually progress to language, reasoning and planning impairments, and there may also be psychiatric symptoms such as hallucinations, behaviour changes such as apathy or agitation, and physical changes such as falls~\cite{orgeta2019lancet,national2007dementia}. Other common causes of dementia include frontotemporal dementia (FTD)~\cite{englund1994clinical,snowden2002frontotemporal}, Lewy body dementia (LBD)~\cite{byrne1995cortical,walker1997neuropsychological} and vascular dementia~\cite{t2015vascular}.

Dementia is typically diagnosed via a specialist doctor who performs a series of clinical assessments including: obtaining a personal, and informant history of the cognitive symptoms, a physical examination, pen and paper cognitive assessments, blood tests to rule out other mimics of dementia (e.g., low vitamin B12 levels), and brain scans to assess for localised brain atrophy. This whole diagnostic process is time consuming, expensive and somewhat subjective - relying heavily on the clinician's interpretation. 

In the last decade, there have been advances in developing new specialist tests to directly detect the pathological proteins of dementia through brain scans and spinal fluid and blood tests; however, these methods remain too invasive, costly or specialist to be widely accessible in clinical practice. There remains an urgent need to find new cost-effective objective and accessible methods to detect AD, and other types of dementia, at a population level. 

Recent developments in computer science, especially Artificial Intelligence (AI), offer a potential solution to this global problem. They provide the technologies that could aid development of new efficient and accessible methods to assist in detecting the earliest stages of AD and other dementias. In this paper, we describe cognitive tests that are currently commonly used and the limitations of these. We discuss AI-based methods that have been applied to analyse cognitive, movement, speech and imaging data for dementia detection, and then outline potential future directions, including smart environments.   

\section{CURRENT COGNITIVE TESTS}
Cognitive tests have been used for decades to screen for, and aid in diagnosis of, dementia. They have formed the mainstay of ‘objective evidence’ of cognitive impairment. However, most of these require clinicians to interpret the results ~\cite{irving1970validity,lorentz2002brief,cukierman2005cognitive,tsoi2015cognitive} and are thus somewhat subjective. They are usually final score based~\cite{velayudhan2014review,tsoi2015cognitive,nasreddine2005montreal} and sensitivity/specificity for MCI and dementia are then defined as per set cut-offs. It is beyond the scope of this paper to describe all such tests but here we set the scene for how AI may improve upon current methods by describing commonly used “quick” screening tests. Table 1 shows the comparative performance of the cognitive screening tests described. 

\subsection{Mini-Mental State Examination (MMSE)}
The MMSE~\cite{folstein1975mini}, first published in 1975, takes 6 to 10 minutes to complete and produces a score out of 30. It measures different aspects of cognition including memory, orientation, language, attention, and visuospatial function. Folstein et al. ~\cite{folstein1975mini}, screened two cohorts for dementia-related diseases, using the MMSE and found significant agreement between MMSE score and dementia diagnosis by clinicians. The mean score for patients with dementia was 9.7, for those with depression and cognitive impairment 19.0, and for controls 27.6. O'Bryant et al.~\cite{o2008detecting} and Crum et al.~\cite{crum1993population} presented the  distribution  of  MMSE  scores by educational and age level and the MMSE has been translated and validated in different regions including Spain~\cite{blesa2001clinical,beaman2004validation}, China~\cite{katzman1988chinese,chiu1994reliability}, France~\cite{gagnon1990validity,park1990modification}, Korea~\cite{han2008adaptation}, Brazil~\cite{kochhann2010mini}, and Greece~\cite{fountoulakis2000mini}. 

The specificity of the MMSE for detecting AD has been calculated as approximately 0.80 - 0.90 but the main drawback is its low sensitivity, especially for detecting early-stage dementia~\cite{lancu2006minimental}. It has proved difficult to set a threshold cutoff score to achieve both high sensitivity and specificity simultaneously. For example, the sensitivity/specificity of the MMSE has been estimated at 0.98/0.32 when a cut-off score of 29 is used, and 0.50/1.00 when a cut-off score is set to 22~\cite{kukull1994mini}.

\subsection{Addenbrooke's Cognitive Examination-Revised (ACE-R)}
The ACE-R is a more extensive cognitive screening test designed by Mioshi et al.~\cite{mioshi2006addenbrooke} and modified from an earlier version of Addenbrooke's Cognitive Examination (ACE) test~\cite{mathuranath2000brief} to include additional visuospatial components. It takes 15-20 minutes to complete, is scored out of 100, and assesses attention, memory, fluency, language and visuospatial function. Nasreddine et al.~\cite{mioshi2006addenbrooke}, applied ACE-R to 142 participants with dementia (67 with AD, 55 with FTD and 20 with LBD), 36 participants with MCI and a control group of 63 participants. The ACE-R achieved a sensitivity/specificity of 0.94/0.89 with a cut-off score of 88, and 0.84/1.00 when the cut-off was set at 82. In addition, ACE-R has been adapted and translated for use in several different countries including Germany~\cite{hsieh2013validation,alexopoulos2010validation}, China~\cite{fang2014validation,wong2013validation}, Japan~\cite{dos2012validation,yoshida2012validation}, Korea~\cite{kwak2010korean}, Brazil~\cite{carvalho2010brazilian}, Spain~\cite{garcia2006validation}, Portugal~\cite{gonccalves2015portuguese}, and France~\cite{bastide2012addenbrooke}. 

\subsection{Montreal Cognitive Assessment (MoCA)}
The MoCA is scored out of 30 points and takes approximately 10 minutes to complete. It tests short-term memory, visuospatial function, executive function, phonemics fluency, attention and language. When Nasreddine et al.~\cite{nasreddine2005montreal} evaluated MoCA in 93 participants with AD, 94 participants with MCI and 90 healthy controls, it had sensitivity/specificity reaching 1.0/0.87 with a cut-off score of 26. Similar to other screening tests, the MoCA has been adapted into various language versions~\cite{lee2008brief,freitas2011montreal,luis2009cross,memoria2013brief,lu2011montreal,yu2012beijing}.

The main drawbacks for all three of these cognitive tests are the need for a clinician, or other trained person, to administer the test and interpret the results, and the time taken to complete these processes. Furthermore, with any test that relies solely on total test score, there is a substantial clinical data about cognitive processes that is simply discarded – for example the recall of a list of objects may be correct (and score full marks) but the delayed time to recall these items, and pauses and hesitations in doing so, is crucial (but unrecorded) information that suggests deficits compared to someone who quickly recalls the same list.   

\subsection{Clock drawing test}
In this deceptively simple and quick test, the person is asked to draw the contour of a clock, write in all the numbers and then draw the clock hands at a particular defined time point (e.g., “ten to five”). The total score for the drawing is based on the accuracy of the contour and the positioning of the numbers and hands and this test assesses executive, spatial and visuo-constructive cognitive functions~\cite{agrell1998clock}. 

\begin{table}[h!]
\caption{COMPARISON OF COMMONLY USED COGNITIVE TESTS}
\label{table}
\centering
\setlength{\tabcolsep}{3pt}
\begin{tabular}{c c c c} % <-- Alignments: 1st column left, 2nd middle and 3rd right, with vertical lines in between
\hline
\textbf{Test} & \textbf{Sensitivity} & \textbf{Specificity} & \textbf{Reference}\\
\hline
MMSE & 79\% & 90\% - 90\% & Folstein et al.~\cite{folstein1975mini} \\
ACE-R & 84\% - 94\% & 89\% & Mioshi et al.~\cite{mioshi2006addenbrooke} \\
MoCA & 84\% - 85\% & 87\% & Nasreddine et al.~\cite{nasreddine2005montreal} \\
Clock drawing test & 85\% & 85\% & Shulman et al.~\cite{shulman2000clock} \\
Clock drawing test & 87\% & 82\% & Watson et al.~\cite{watson1993clock} \\
\hline
\end{tabular}
\end{table}

Shulman et al.~\cite{shulman2000clock} reviewed clock drawing tests and concluded that mean sensitivity/specificity was approximately 0.85/0.85. Watson et al’s.~\cite{watson1993clock} adapted clock drawing test requires participants to draw clock numbers in a pre-drawn circle and the number of digits in each quadrant is counted to calculate the error, achieving a sensitivity/specificity of 0.87/0.82 for dementia. Although clock drawing tests are relatively quick to administer and achieve similar sensitivity and specificity to other longer cognitive screening tests, the quality of the clock drawn by participants has not been appropriately considered in most cases. For example, there are likely to be subtle differences during the process of drawing, such as slowing down, pausing, hesitations, etc. that signal early changes of dementia, but all this data is disregarded in the simple total score analysis.  

In summary, the limitations or the current methods for screening for cognitive deficits associated with dementia are the time taken, the need for a trained person to administer and interpret the test, the reduction of performance to a total score that loses crucial information on cognitive decline, and the lack of sensitivity to the earliest stages of dementia. These same issues also apply to detailed ‘gold standard’ neuropsychological tests to a greater or lesser degree, although the trade-off for greater sensitivity/specificity is also greater time consumption and need for an expert to administer. There thus remains a need for new automated, objective, accurate and cost-effective methods that analyse not only the outcome at the end of testing but also cognitive performance \textit{during} the process.

\section{Artificial Intelligence (AI) Based Tests}
Recently, various AI-based tests have been introduced to assist with the detection of dementia – either through detecting functional changes (such as cognitive, movement or speech impairments) or through detecting pathological abnormalities on brain scans. AI-based tests can provide improved accuracy through their capability of capturing additional features from a large amount of data. This brings out more objective inference compared with clinicians' manually analyzed results~\cite{danso2019application}. Furthermore, AI provides an automated analysis process - both in terms of time and cost efficiencies~\cite{williams2013machine}. Recently introduced AI-based tests include computerized cognitive tests, computer-assisted interpretation of brain scans, and movement – and speech-analysis tests. None of these have yet reached routine clinical practice and largely remain in the research setting. Here we discuss the techniques used and the relative accuracies of the various approaches.

\subsection{Computerized Cognitive Tests}
In recent years, computerized cognitive tests have been introduced for screening and detection of dementia~\cite{angelillo2019attentional,thabtah2020mobile}. They collect data on participants' behaviour during the tests as well as their total scores. Other benefits include the potential to be user-directed, negating the need for a researcher or clinician to administer it - thus reducing costs, improving accessibility, and avoiding inter-rater variability.

CogState is a computerized, web-based battery that measures visuomotor function, psychomotor speed, attention, memory, execution function and social cognition~\cite{lim2012use}. It is easy to understand for testers having less computer using experience. It also provides customized batteries which are subsets of the general CogState battery to satisfy customized requirements. Several works examine the validity of CogState battery. Maruff et al.~\cite{maruff2009validity} found that CogState measures had strong correlations with conventional neuropsychological measures with correlation coefficient ranging from 0.49 to 0.83. Mielke et al. validated the performance of the CogState battery in the Mayo Clinic Study on Aging and found the correlations between CogState and neuropsychological tests were ranging from -0.46 to 0.53~\cite{mielke2015performance}.

CANTAB battery, first developed in 1992~\cite{sahakian1992computerized}, has become a computerized battery that has been validated by 30 years of global neuroscience research~\cite{barnett2015paired}. It consists of working memory, executive function, visual, verbal memory, attention, information processing time, social and emotion recognition, decision making and response control. 

Based on MoCA, Yu et al.~\cite{yu2015development} developed a MoCA-CC (computerized MoCA), which is a person-to-machine pattern with input system, test system, output system, and management system, to screen MCI and normal. MoCA-CC achieved a sensitivity/specificity of 0.96/0.87.

Memoria et al.~\cite{memoria2014contributions} et al. developed a Computer-Administered Neuropsychological Screen for Mild Cognitive Impairment (CANS-MCI), which integrated the administration and scoring module and can provide immediate analysis. CANS-MCI was applied to discriminate AD, MCI and normal groups and achieved a sensitivity/specificity of 0.81/0.73 for normal vs MCI and 1.00/0.97 for normal and AD.

Computer-based clock drawing tests have been developed through machine learning methods, such as supporting vector machine (SVM), logistic regression and random forest. Angelillo et al.~\cite{angelillo2019attentional} designed a digitized Attentional Matrices Test (AMT) that involved participants drawing a clock on a digitizing tablet. They used an ensemble method by evaluating the performance of four machine learning methods (i.e. K-Nearest Neighbours (KNN), logistic regression model, SVM and random forest) to classify subjects into dementia and healthy control groups and achieved a sensitivity/specificity of 0.86/0.83. Shigemori et al.~\cite{shigemori2016dementia} designed a similar tablet-based Clock Drawing Test (CDT); they used SVM to classify subjects into Vascular Dementia (VaD), MCI and AD with a sensitivity/specificity of 0.97/0.97. Bennasar et al.~\cite{bennasar2014cascade} focused on analyzing geometric data derived from the computerized clock drawing test, such as degrees of angles, were extracted to discriminate between MCI, severe/moderate dementia and normal. A total of 47 clock features combined with MMSE data were fed into three cascaded classifiers for classification. The classifier achieved an accuracy of 77.8\% in classifying the three cases.

With the help of machine learning technologies, additional features can be extracted, which improves the speed and accuracy of the screening process, but the main drawback is the requirement for a tablet - this may affect its usability and accessibility. Table 2 shows the performance of different computerized cognitive screening tests.

\begin{table}[h!]
\caption{Computerized cognitive screening test.}
\label{table}
\centering
\setlength{\tabcolsep}{3pt}
\begin{tabular}{c c c c}
\hline
\textbf{Method} & \textbf{Test} & \textbf{Sensitivity} & \textbf{Specificity}\\
\hline
Ensemble ML method & Digitized Attentional Matrices Test~\cite{angelillo2019attentional} &  86.11\% & 82.76\% \\
SVM & Tablet based Clock Drawing Test~\cite{shigemori2016dementia} & 97\% & 97\% \\
Random forest & Computerized handwriting task~\cite{impedovo2019handwriting} & 72\% & 73\%  \\
Cascaded ML classifier & Computerized Clock Drawing Test~\cite{bennasar2014cascade} & 89\% & No Info. \\
\hline
\end{tabular}
\end{table}

\subsection{Movement Tests}
Another emerging approach for screening dementia is to analyse patterns of voluntary movement as motor control is known to deteriorate in dementia~\cite{lerche2016prospective,buracchio2010trajectory,aggarwal2006motor,boyle2005parkinsonian,wilson2003parkinsonianlike,camicioli1998motor}. Gait~\cite{beauchet2016association,buckley2019role}, hand movements~\cite{mollica2019early}, and eye movements are the three main types of movement that have been investigated.

Riona et al.~\cite{mc2020differentiating} tried to differentiate AD, DLB and PDD through gait analysis by using a accelerometer-based wearable. Participants were required to perform six 10m walks at their comfortable pace with the wearable setting on the skin above the fifth lumbar vertabra. Fourteen gait features representing pace, variability, rhythm, asymmetry and postural control have been extracted to discriminate three cases. According to the AUC metric, only variability and asymmetry features can significantly discriminate AD from PDD. Chung et al.\cite{chung2012gait} set similar walking task for 12 participants (9 AD and 3 healthy controls) to do the gait analysis. Participants were required to walk along straight line of 40m without speed limit and a triaxial accelerometer was set on the foot to extract gait features. The result showed different spatiotemporal gait parameters between AD and healthy controls but without statistical analysis. Few studies have used machine learning methods to analyse gait. Beauchet et al.~\cite{beauchet2016association} found that gait analysis could differentiate between motoric cognitive risk (MCR), a pre-dementia syndrome characterized by slow gait and cognitive complaints~\cite{annweiler29motoric}, healthy people and people with dementia. In their experiment, participants were divided into MCR class and non-MCR class. Gait speed combined with other data including participant's cognitive test results, multiple logistic regression model and statistical tests were used to determine the relationship between MCR syndrome and brain volumes. The results showed that for executive function, there was significant difference between MCR and non-MCR groups.

Motor control of the hands and dexterity is also known to change in dementia and AI methods have been applied in a research setting. For example, Sano et al.~\cite{sano2019detection} used a magnetic sensor-based device to measure participants' finger-tapping waveforms and extracted 15 features for one-class SVM. The model achieved a recall of over 95\% to classify normal subjects from those with dementia. Liang et al.~\cite{liang2019real} investigated two hand trajectory tracking methods to assist in screening deaf individuals for signs of dementia. The first method was based on skin colour segmentation, and the second method was based on part affinity fields OpenPose Skeleton Model. According to the comparison experiments, the second method showed more enhanced and robust performance on tracking hand trajectory in terms of the distance between prediction trajectory and ground truth trajectory in a coordinate system.

Computer-assisted graphic drawings and handwriting tests are an alternative method to examine hand motor function. Yu et al.~\cite{yu2019characterization} designed several graphic drawing and handwriting tasks to assess the differences in motor function between AD, MCI and healthy controls. Movement fluency features (such as pause time per stroke and the ratio of in air to on paper time) and handwriting accuracy related features, such as character size variation and stroke orientation control, were extracted. These features were combined with personal information compared between the groups. The result showed that AD and MCI groups had significantly larger variation and errors than healthy subjects in terms of the control of line drawing. Impedovo et al.~\cite{impedovo2019handwriting} designed a protocol which integrated caregiver interview, a traditional cognitive test (MMSE) and a set of computerized handwriting tasks. Handwriting features including time stamp, x-y coordinates, and mean pressure were also extracted and fed into the random forest classifier to achieve an average precision of 0.72, recall of 0.73 and F-measure of 0.71.

Eye movement tests have also been explored as a non-invasive technique for the early detection of dementia~\cite{crutcher2009eye,anderson2013eye,chatterjee2018eye,currie1988eye,beltran2018computational}. For example, in~\cite{fraser2017analysis} healthy people, people with subjective cognitive impairment (SCI) and people with MCI read aloud and then read silently whilst their eye movements were measured by EyeLink 1000 Desktop Mount with monocular eye-tracking and a headrest for head stabilization. Thirteen eye movement features including saccade amplitude and total fixations were extracted, and then machine learning-based classifiers such as Naive Bayes model, SVM and logistic regression (LR) model were applied. The Naive Bayes model achieved the best accuracy of 86.0\% to discriminate MCI, SCI and normal. In another study, and eye-tracking system was proposed to monitor 15 participants' (9 older controls, 3 young controls and 3 MCI) cognitive function through a memory test and an eye movement task. Eye movement task required participants to watch four short videos with eyes fixating on one object during the video. The same 13 features were extracted as in~\cite{fraser2017analysis}, and fed into least square regression, ridge regression and LASSO regression to perform the classification. The LASSO regression model achieved the highest accuracy to predict memory ability based on eye movement with a mean absolute error of 5.52~\cite{zhang2016monitoring}.

Unlike most cognitive tests, movement-based tests offer the opportunity to examine for declining brain functions using a method free of language. This is an attractive feature when considering accessibility of tests to populations around the world. However, this field of research is still somewhat in its infancy and constrained by the need generally for additional equipment to measure the body movements accurately. It may be that 'smart' environments solve this issue in the future (See Section IV below).

\subsection{Speech, Conversation, and Language Tests}
Speech and language abilities are known to be impaired early in several types of dementia, with symptoms such as aphasia, pauses, reduced vocabulary and other language impairments~\cite{hopper2007service,kempler2008language,snowden2002frontotemporal}. Several studies have used AI-based speech and language tests for dementia screening. Common test processes include extracting related features and then feeding the features into machine learning or deep learning classifiers to detect the patterns consistent with dementia. There are mainly two types of features that can be extracted and analyzed, i.e., acoustic features and linguistic features. Acoustic features describe how people articulate speech - the actual sounds rather than the content. Common acoustic features include Mel-frequency cepstral coefficients (MFCC), speech rate, and cepstral peak prominence (CPP). Linguistic features describe the content, such as the vocabulary, grammar and syntax. Classical machine learning methods, inclusive of SVM, logistic regression model, random forest and KNN, have been applied to classify subjects into different categories based on the extracted features. Two main types of speech tests have been used: a picture description test and conversation, generated through an interview test.

For picture description tests ~\cite{haider2019assessment,chien2019automatic,liu2019dementia,pan2019automatic}, participants are asked to describe a picture in a fixed time and speech is recorded. Liu et al.~\cite{liu2019dementia} used this method and analyzed the spontaneous speech for screening dementia and MCI. Temporal features (e.g., duration of each utterance, number of syllables), acoustic features (e.g., energy, jitter and shimmer, MFCCs) and linguistic features (e.g., word counts) were extracted and then fed into a logistic regression classifier, achieving an accuracy of 81.9\%. Likewise, Haider et al.~\cite{haider2019assessment} evaluated the performance of a model trained by purely acoustic features for detecting dementia. Data were collected from four speech tasks, inclusive of a description task, a word fluency task, a story recall task and a sentence construction task. Extracted acoustic features included eGeMAPS, emobase, ComParE and MRCG. Based on acoustic features, they used active data representation (ADR) to capture all speech segments information. KNN, linear discriminant analysis (LDA), random forests and SVM were used in classification experiments for dementia detection. The highest classification accuracy, of 77.4\%,  was achieved by using LDA on eGeMAPS features. 

In conclusion, it has been shown that the data features of speech have a greater influence on accuracy than the machine learning method. Combining both linguistic features and acoustic features performs better than each on its own. Pan et al.~\cite{pan2019automatic} proposed an automatic hierarchical bidirectional attention neural network to detect AD based on a picture description task and their model achieved an accuracy of 84.0\%. Chien et al.~\cite{chien2019automatic} used a convolutional recurrent neural network (CRNN) to generate feature sequences. The network achieved a sensitivity/specificity  of 0.76/0.76. In conclusion, deep learning techniques did not obviously show many advantages over machine learning in terms of classification accuracy. One possible reason can be the lack of data for deep learning method training.

The second approach to speech analysis for dementia detection is to analyse conversational speech generated through interviews~\cite{tanaka2017detecting,mirheidari2018detecting,luz2018method,ujiro2018detection,chien2018assessment,weiner2018investigating,mirheidari2019dementia}. Mirheidari et al.~\cite{mirheidari2018detecting} used word2vector and GloVe methods to extract natural language-related features from participants' conversations. They achieved approximately 60\% accuracy on average across a range of datasets. Mirheidari et al. continued their works~\cite{mirheidari2019dementia} with the help of SHoUT diarization toolkit and automatic speech recognition (ASR) for extracting 12 acoustic features, 12 lexical features and 20 conversation analysis (CA)-based features based on interviews between patients and neurologists. The classification accuracy on discriminating between controls and FMD reached 90.0\%. Weiner et al.~\cite{weiner2017manual} proposed a framework to detect early signs of dementia based on conversations, consisting of voice activity detection, speaker diarization, feature extraction and dementia detection. Speech pause-based features, speaking rate features, word error rate features and linguistic features from spoken dialogues through a computer avatar were extracted and fed to SVM and logistic regression models to distinguish Aging-Associated Cognitive Decline (AACD), AD and healthy controls. Tanaka et al.~\cite{tanaka2017detecting}, Ujiro et al.~\cite{ujiro2018detection} and Luz et al.~\cite{luz2018method} were similar works but used subjects' spoken dialogues with doctors to perform the analysis and found 86.5\% - 95.0\% classification accuracy for controls vs dementia. Bullard et al.~\cite{bullard2016towards}, Shao et al.~\cite{shao2019detection} and Balbim et al.~\cite{balbim2020evaluation} attempted to combine speech, conversation and language data with other sources of data to improve detection performance. For example, Bullard et al.~\cite{bullard2016towards} fused linguistic data, such as bag-of-words (BOW), term frequency inverse document frequency (TF-IDF) and topics from latent Dirichlet allocation (LDA), with non-linguistic clinical data such as blood tests and cerebrospinal fluid biomarkers. These multi-modal features were fed into a logistic regression to classify subjects into AD, Late MCI, Early MCI and healthy control groups. The highest (MCI, AD and normal) classification accuracy reached 80.9\%. Table 3 shows the comparative performance of computerized speech analysis tests.

\begin{table}[h!]
\caption{Performance of computerized speech analysis tests to discriminate dementia from controls.}
\label{table}
\centering
\setlength{\tabcolsep}{3pt}
\begin{tabular}{c c c}
\hline
\textbf{Method} & \textbf{Test} & \textbf{Accuracy} \\
\hline
Logistic regression &  Spontaneous speech test~\cite{liu2019dementia} & 81.9\% \\
SVM & 4 different types of speech~\cite{haider2019assessment} & 77.4\% \\
Attention neural network & Picture description test~\cite{pan2019automatic} & 84.02\% \\
Machine learning method & Conversation~\cite{mirheidari2018detecting} & 60\% \\
Machine learning method & Conversation~\cite{mirheidari2019dementia} & 90\% \\
Bag-of-words (BOW), logistic regression & Conversation~\cite{bullard2016towards} & 80.9\% \\
Convolutional recurrent neural network & Conversation~\cite{chien2019automatic} & 75.6\% \\      
\hline
\end{tabular}
\end{table}

\subsection{Computer - Assisted Interpretation of Brain Scans}
Currently, magnetic resonance imaging (MRI) and computed tomography (CT) scans are used to support the differential diagnosis that is formed though cognitive tests and clinical assessment. Brain scans provide information on structural changes of various brain regions that are known to be associated with different types of dementia. For example, clinicians will specifically look for atrophy (shrinkage) in a part of the brain called the hippocampus in AD, or in the frontal lobes in FTD. Positron emission tomography (PET) is a specialist and expensive brain scan, only accessible in certain centres, that provides information on function (blood flow and metabolism of brain regions) and can also measure underlying abnormal proteins such as amyloid (in AD). The presence of these changes provides additional evidence to support the clinical assessment, but the interpretation of the scans is somewhat subjective and vulnerable to inter-rater variability. Furthermore, specialist doctors are required to interpret the visual data of a scan – a precious resource that many people around the world do not have access to. Early-stage dementia may have only minimal, or no atrophy and so is often the most challenging scan to interpret as it sits on the cusp of normality. A computerized method to aid automated interpretation and measurement of brain scans would be highly advantageous to the process of diagnosing dementia.

Two main AI methods have been used to analyse brain scans: machine learning and deep learning. A recent systemic review~\cite{pellegrini2018machine} examined machine learning methods to classify subjects based on neuroimaging data.

Various machine learning models including AdaBoost, Gaussian process, logistic regression, random forest~\cite{mathotaarachchi2017identifying}, support vector machine (SVM), and k nearest neighbour (kNN), have been applied to two large longitudinal data sets that together comprise interval MRI and PET scans from almost 2,000 participants with dementia, MCI or healthy controls: Alzheimer's Disease Neuroimaging Initiative (ADNI)~\cite{mueller2005alzheimer} and Open Access Series of Imaging Studies (OASIS)~\cite{marcus2007open}. For example, Mathotaarachchi et al.~\cite{mathotaarachchi2017identifying} focused on methods to predict the progression of MCI to dementia using ADNI image data. They pre-processed PET data using CIVET pipeline and the outputs of the pipeline were Florbetapir PET Standard Update Value Ratios (SUVR). Based on SUVR values, features were extracted by applying voxel-wise logistic regression analysis and a random forest-based classifier was used for training and prediction. The classifier achieved a sensitivity/specificity of 0.71/0.87 to determine MCI from dementia. Although machine learning based brain scan analysis has shown promising results in some studies, Pellegrini et al.~\cite{pellegrini2018machine} stated that generally it has proved hard to discriminate the early imaging changes of dementia with machine learning methods.

Recently, deep learning techniques have provided more promising results for detecting the subtle imaging changes associated with the earliest stages of dementia~\cite{bidani2019dementia,iizuka2019deep,islam2018early}. Yiugit et al.~\cite{yiugit2020applying} used a CNN model to classify MRI scans from people with AD, MCI and healthy controls. Slices from the 3D MRI scans were extracted, data augmentation was performed, then three different neural network models with distinct numbers of convolutional layers were compared. The proposed method achieved a sensitivity/specificity of 0.93/0.81 to discriminate MCI from normal and 0.86/0.75 to discriminate AD from normal. There are also encouraging results that deep learning can differentiate different typs of dementia; for example Iizuka et al.~\cite{iizuka2019deep} used a deep CNN model trained by brain PET scan data to distinguish DLB, AD and healthy controls; their model had a binary classification accuracy of 93.1\% between DLB and controls, 89.3\% between DLB and AD and 92.4\% between AD and controls. Backstorm et al.~\cite{backstrom2018efficient} applied 3D CNNs on 3D MRI brain scans to classify healthy controls from individuals with AD. The model achieved an accuracy of 98.74\%, with an AD detection rate of 100\% and a false positive rate of 2.4\%. 

In summary, deep learning methods show promising results for accurately discriminating brain scans indicative of AD compared to healthy ageing. They also show promise to discriminate specific types of dementia. Rather than depending solely on clinicians' subjective judgement, deep learning approaches have strong potential to provide automated, objective measures and reduce the inter- and intra- subject variability. However, current challenges are that deep learning models require a huge amount of data for training and validating purposes. Furthermore, CNN-based models lack interpretability, to understand which precise features of the scan they base their models on. This “black box” approach may hinder acceptance by clinicians and integration into standard practice. Table 4 shows the computer-assisted method performance in analyzing medical scan images for detecting changes associated with dementia. 

\begin{table}[h!]
\caption{Computer-assisted method performance on medical scan tests.}
\label{table}
\centering
\setlength{\tabcolsep}{3pt}
\begin{tabular}{c c c c c}
\hline
\textbf{Method} & \textbf{Sensitivity} & \textbf{Specificity} & \textbf{Dataset} & \textbf{Reference}\\
\hline
Random forest & 70.8\% & 86.5\% & ADNI & Mathotaarachchi et al.~\cite{mathotaarachchi2017identifying} \\
CNN-based & 93\% & 81\% & MRI data & Yiugit et al.~\cite{yiugit2020applying} \\
CNN-based & 92.4\% & No Info. & OASIS & Iizuka et al.~\cite{iizuka2019deep} \\
3D CNNs & 98.74\% & No Info. & MRI data & Backstorm et al.~\cite{backstrom2018efficient} \\
3D CNNs & 85\% & No Info. & MRI data & Kompanek et al.~\cite{kompanek2019volumetrie} \\
\hline
\end{tabular}
\end{table}

\section{Future directions: combining AI methods and smart environments}
Smart environments and combinations of AI-based methods are exciting potential future research directions. Establishing smart environments is a novel approach that, in the future, could help detect changes of early dementia in people's own homes~\cite{lotfi2012smart}. The so-called “smart” environment comprises different “intelligent”, or AI-driven, sensors surrounding the living environment to monitor daily activities without imposing restrictions compared to using wearable sensors. There are two main categories of smart environment monitoring, i.e. real-life monitoring and scenario-based monitoring. Real-life monitoring refers to the situation that participants simply go about their usual daily activities with different sensors being installed around their home, while scenario-based monitoring refers to participants being asked to enter a research laboratory to simulate the real-life situation.

Batista~\cite{batista2016wandering} reviewed the detection of wandering patterns (a sign of early dementia) through a smart environment. Trajectory analysis was based on the position data, which were captured by GPS, Infra-Red and RFID proximity sensors. The movement data was fed into machine learning methods to classify wandering, from non-wandering patterns. They concluded that, due to the uncertainty of the environment, and uniqueness of individual behaviour patterns, detecting wandering behaviours in a smart environment is still a challenge. Khodabandehloo et al.~\cite{khodabandehloo2020collaborative} set ambient sensors to detect wandering behaviour by monitoring people's real-life activities. They adopted a collaborative learning technology to complete two tasks, i.e. trajectory segmentation and wandering episode detection. A series of position records captured by ambient sensors were used for trajectory segmentation, and wandering episodes were detected based on the trajectory segmentation by using machine learning methods. Among a range of machine learning methods, the random forest model achieved the highest overall accuracy of 80.7\% for the case of personalized selection and 70.5\% for the case of non-personalized selection.

In addition to wandering pattern detection, movement analysis in smart environments has also assisted detecting early signs of cognitive related disorders. Paolini et al.~\cite{paolini2017human} set a smart environment for early detection of AD based on localization and movement analysis. RFID reader technology was used for indoor localization analysis, and a wearable inertial sensor system including accelerometer and gyroscope was used for movement analysis. Early detection of AD was further realized by evaluating indoor movement activities. Stavropoulos et al.~\cite{stavropoulos2016semantic} introduced a framework by integrating different ambient sensors’ data to analyze movement and activities, which supported the care for people with dementia. Sensors included a camera, microphone, DTI-2 (a wearable device), and plugs, which collected image data, audio data, and motion data. They set different sensors in a room where a person with dementia lives. By using different types of sensors, different types of data could be extracted to monitor the person. Such heterogeneous data collected from different sensors were being processed at an analysis layer, then went to a representation layer to store the knowledge. After that, it went to an interpretation layer for semantic interpretation and fusion. Furthermore, it went to a service layer and an application layer for clinicians to use. For visual activity recognition, they used GMM clustering method to cluster specific activities. Fisher vectors were used for extracting fixed size features. Multiclass SVM model was used for the classification of different activities. On a dataset of 98 participants, the model achieved an average recall and precision of 82\% for recognition of four activities (answering a phone, establishing account balance, preparing a drink, preparing drug box).

Similarly, Lotfi et al.~\cite{lotfi2012smart} set passive infra-red sensors (PIR) around the home, such as door entry-point sensors, electricity power usage sensors, bed/sofa pressure sensors and flood sensors, to predict subjects' abnormal behaviour through movement analysis. Echo state network (ESN) was used to classify different behaviours. The root mean square error (RMSE) of training was around 1\% and 7\% for back door and kitchen sensors respectively.

In conclusion, using smart home environments to capture daily life behaviours may provide a completely new method in the future to detect the earliest changes associated with dementia. However, the complex environment and the diversity of events lead to more challenges. Privacy and acceptability to the public are other issues that need to be explored.

\section{Conclusions}
There is a growing field of research that applies AI technologies to aid the early detection of dementia screening tests. AI-based tests offer exciting opportunities to streamline the diagnostic process of dementia detection. They may assist clinicians by providing objective evidence of changes associated with the early stages of dementia and offer several advantages over traditional dementia screening tests. First, AI enables more features to be retrieved from a single test, which may improve the accuracy of dementia detection. For example, compared with traditional score-based cognitive tests, computerized cognitive tests can not only extract the final test results but also the richer process-related features. Second, AI helps reduce errors due to subjective judgements. For example, how a patient performs in a conversation test is largely dependent on the interviewee’s subjective opinion at that time but AI makes the inference process more objective. Third, AI shifts the automation of dementia screening to a higher level and this is particularly attractive to epidemiology studies and to public health bodies who want to target risk reduction interventions at an early stage. AI is quick to analyze and respond to large population screening compared with doctors’ judgement alone.

We propose two main future directions for AI-based dementia screening: smart environments and non-invasive multiple tests integration. The smart environment can monitor candidates’ behaviour in various aspects of daily life by different environmental sensors. Current smart environment applications generally assess participants whilst they perform specific tasks or tests. In the future, the smart environment could play a greater role in monitoring people’s behaviour in normal daily life and recognise personalized change over time offering potential for a precision approach to detection and monitoring of dementia.

Another key future direction will be overcoming the challenge of how to integrate different dementia screening tests – to make the multi-modal tool as accurate as possible. AI-based dementia tests so far have been designed to work separately. For example, language tests are designed to screen dementia in terms of language abilities, while cognitive tests focus on functions like memory and attention. It would be particularly synergistic to see speech and motion analysis combined in smart environments, especially as we know that dual motor-speech tasks tend to be impaired early in dementia~\cite{poole2017motor}. Furthermore, a strength of AI, especially deep learning, is integrating results from different tests to find the most discriminating patterns.

The challenges in the design and the implementation processes AI-based dementia screening tests are critical. For example, how do we design a comprehensive, easily accessible, and swift protocol that integrates different types of data to provide accurate detection of dementia? Perhaps the hardest challenge of all will be opening up the black box models of AI so that providers, clinicians and users alike feel they can trust the outcomes and understand the findings.

\bibliographystyle{unsrt}  
\bibliography{references.bib}

\begin{thebibliography}{100}

\bibitem{livingston2017dementia}
Gill Livingston, Andrew Sommerlad, Vasiliki Orgeta, Sergi~G Costafreda,
  Jonathan Huntley, David Ames, Clive Ballard, Sube Banerjee, Alistair Burns,
  Jiska Cohen-Mansfield, et~al.
\newblock Dementia prevention, intervention, and care.
\newblock {\em The Lancet}, 390(10113):2673--2734, 2017.

\bibitem{world2019risk}
World~Health Organization et~al.
\newblock Risk reduction of cognitive decline and dementia: Who guidelines.
\newblock 2019.

\bibitem{livingston2020dementia}
Gill Livingston, Jonathan Huntley, Andrew Sommerlad, David Ames, Clive Ballard,
  Sube Banerjee, Carol Brayne, Alistair Burns, Jiska Cohen-Mansfield, Claudia
  Cooper, et~al.
\newblock Dementia prevention, intervention, and care: 2020 report of the
  lancet commission.
\newblock {\em The Lancet}, 396(10248):413--446, 2020.

\bibitem{orgeta2019lancet}
Vasiliki Orgeta, N~Mukadam, A~Sommerlad, and G~Livingston.
\newblock The lancet commission on dementia prevention, intervention, and care:
  a call for action.
\newblock {\em Irish journal of psychological medicine}, 36(2):85--88, 2019.

\bibitem{national2007dementia}
National Collaborating~Centre for Mental Health~(UK et~al.
\newblock Dementia.
\newblock In {\em Dementia: A NICE-SCIE Guideline on Supporting People With
  Dementia and Their Carers in Health and Social Care}. British Psychological
  Society, 2007.

\bibitem{englund1994clinical}
B~Englund, A~Brun, L~Gustafson, U~Passant, D~Mann, D~Neary, and JS~Snowden.
\newblock Clinical and neuropathological criteria for frontotemporal dementia.
\newblock {\em J Neurol Neurosurg Psychiatry}, 57(4):416--8, 1994.

\bibitem{snowden2002frontotemporal}
Julie~S Snowden, David Neary, and David~MA Mann.
\newblock Frontotemporal dementia.
\newblock {\em The British journal of psychiatry}, 180(2):140--143, 2002.

\bibitem{byrne1995cortical}
E~JANE Byrne, R~Levy, and R~Howard.
\newblock Cortical lewy body disease: an alternative view.
\newblock {\em Levy R, et al. Developments in dementia and functional disorders
  in the elderly. Petersfield, UK: Wrightson Biomedical}, pages 21--30, 1995.

\bibitem{walker1997neuropsychological}
Zuzana Walker, Ruth~L Allen, Sukhwinder Shergill, and Cornelius~LE Katona.
\newblock Neuropsychological performance in lewy body dementia and alzheimer's
  disease.
\newblock {\em The British Journal of Psychiatry}, 170:156, 1997.

\bibitem{t2015vascular}
John T~O'Brien and Alan Thomas.
\newblock Vascular dementia.
\newblock {\em The Lancet}, 386(10004):1698--1706, 2015.

\bibitem{irving1970validity}
G~Irving, RA~Robinson, and W~McAdam.
\newblock The validity of some cognitive tests in the diagnosis of dementia.
\newblock {\em The British Journal of Psychiatry}, 117(537):149--156, 1970.

\bibitem{lorentz2002brief}
Wendy~J Lorentz, James~M Scanlan, and Soo Borson.
\newblock Brief screening tests for dementia.
\newblock {\em The Canadian Journal of Psychiatry}, 47(8):723--733, 2002.

\bibitem{cukierman2005cognitive}
T~Cukierman, HC~Gerstein, and JD~Williamson.
\newblock Cognitive decline and dementia in diabetes—systematic overview of
  prospective observational studies.
\newblock {\em Diabetologia}, 48(12):2460--2469, 2005.

\bibitem{tsoi2015cognitive}
Kelvin~KF Tsoi, Joyce~YC Chan, Hoyee~W Hirai, Samuel~YS Wong, and Timothy~CY
  Kwok.
\newblock Cognitive tests to detect dementia: a systematic review and
  meta-analysis.
\newblock {\em JAMA internal medicine}, 175(9):1450--1458, 2015.

\bibitem{velayudhan2014review}
Latha Velayudhan, Seung-Ho Ryu, Malgorzata Raczek, Michael Philpot, James
  Lindesay, Matthew Critchfield, and Gill Livingston.
\newblock Review of brief cognitive tests for patients with suspected dementia.
\newblock {\em International psychogeriatrics}, 26(8):1247--1262, 2014.

\bibitem{nasreddine2005montreal}
Ziad~S Nasreddine, Natalie~A Phillips, Val{\'e}rie B{\'e}dirian, Simon
  Charbonneau, Victor Whitehead, Isabelle Collin, Jeffrey~L Cummings, and
  Howard Chertkow.
\newblock The montreal cognitive assessment, moca: a brief screening tool for
  mild cognitive impairment.
\newblock {\em Journal of the American Geriatrics Society}, 53(4):695--699,
  2005.

\bibitem{folstein1975mini}
Marshal~F Folstein, Susan~E Folstein, and Paul~R McHugh.
\newblock “mini-mental state”: a practical method for grading the cognitive
  state of patients for the clinician.
\newblock {\em Journal of psychiatric research}, 12(3):189--198, 1975.

\bibitem{o2008detecting}
Sid~E O’Bryant, Joy~D Humphreys, Glenn~E Smith, Robert~J Ivnik, Neill~R
  Graff-Radford, Ronald~C Petersen, and John~A Lucas.
\newblock Detecting dementia with the mini-mental state examination in highly
  educated individuals.
\newblock {\em Archives of neurology}, 65(7):963--967, 2008.

\bibitem{crum1993population}
Rosa~M Crum, James~C Anthony, Susan~S Bassett, and Marshal~F Folstein.
\newblock Population-based norms for the mini-mental state examination by age
  and educational level.
\newblock {\em Jama}, 269(18):2386--2391, 1993.

\bibitem{blesa2001clinical}
Rafael Blesa, Montse Pujol, Miguel Aguilar, Pilar Santacruz, Imma
  Bertran-Serra, Gonzalo Hern{\'a}ndez, Jos{\'e}~M Sol, Jordi
  Pe{\~n}a-Casanova, NORMACODEM Group, et~al.
\newblock Clinical validity of the ‘mini-mental state’for spanish speaking
  communities.
\newblock {\em Neuropsychologia}, 39(11):1150--1157, 2001.

\bibitem{beaman2004validation}
Sandra Reyes~de Beaman, Peter~E Beaman, Carmen Garcia-Pena, Miguel~Angel Villa,
  Julieta Heres, Alejandro C{\'o}rdova, and Carol Jagger.
\newblock Validation of a modified version of the mini-mental state examination
  (mmse) in spanish.
\newblock {\em Aging, Neuropsychology, and Cognition}, 11(1):1--11, 2004.

\bibitem{katzman1988chinese}
Robert Katzman, Mingyuan Zhang, Zhengyu Wang, William~T Liu, Elena Yu, Sin-Chi
  Wong, David~P Salmon, Igor Grant, et~al.
\newblock A chinese version of the mini-mental state examination; impact of
  illiteracy in a shanghai dementia survey.
\newblock {\em Journal of clinical epidemiology}, 41(10):971--978, 1988.

\bibitem{chiu1994reliability}
Helen~FK Chiu, HC~Lee, WS~Chung, and PK~Kwong.
\newblock Reliability and validity of the cantonese version of mini-mental
  state examination-a preliminary study.
\newblock {\em Hong Kong Journal of Psychiatry}, 4(2):25, 1994.

\bibitem{gagnon1990validity}
Mich{\`e}le Gagnon, Luc Letenneur, Jean-Fran{\c{c}}ois Dartigues, Daniel
  Commenges, Jean-Marc Orgogozo, Pascale Barberger-Gateau, Annick
  Alp{\'e}rovitch, Arnaud D{\'e}camps, and Roger Salamon.
\newblock Validity of the mini-mental state examination as a screening
  instrument for cognitive impairment and dementia in french elderly community
  residents.
\newblock {\em Neuroepidemiology}, 9(3):143--150, 1990.

\bibitem{park1990modification}
Jong-Han Park and Yong~Chul Kwon.
\newblock Modification of the mini-mental state examination for use in the
  elderly in a non-western society. part 1. development of korean version of
  mini-mental state examination.
\newblock {\em International Journal of Geriatric Psychiatry}, 5(6):381--387,
  1990.

\bibitem{han2008adaptation}
Changsu Han, Sangmee~Ahn Jo, Inho Jo, Eunkyung Kim, Moon~Ho Park, and Yeonwook
  Kang.
\newblock An adaptation of the korean mini-mental state examination (k-mmse) in
  elderly koreans: demographic influence and population-based norms (the age
  study).
\newblock {\em Archives of gerontology and geriatrics}, 47(3):302--310, 2008.

\bibitem{kochhann2010mini}
Renata Kochhann, Juliana~Santos Varela, Carolina~Saraiva de~Macedo~Lisboa, and
  M{\'a}rcia Lorena~Fagundes Chaves.
\newblock The mini mental state examination: review of cutoff points adjusted
  for schooling in a large southern brazilian sample.
\newblock {\em Dementia \& Neuropsychologia}, 4(1):35, 2010.

\bibitem{fountoulakis2000mini}
Konstantinos~N Fountoulakis, Magda Tsolaki, Helen Chantzi, and Aristides Kazis.
\newblock Mini mental state examination (mmse): a validation study in greece.
\newblock {\em American Journal of Alzheimer's Disease \& Other
  Dementias{\textregistered}}, 15(6):342--345, 2000.

\bibitem{lancu2006minimental}
Iulian Lancu and Ahikam Olmer.
\newblock The minimental state examination--an up-to-date review.
\newblock {\em Harefuah}, 145(9):687--701, 2006.

\bibitem{kukull1994mini}
WA~Kukull, EB~Larson, L~Teri, J~Bowen, W~McCormick, and ML~Pfanschmidt.
\newblock The mini-mental state examination score and the clinical diagnosis of
  dementia.
\newblock {\em Journal of clinical epidemiology}, 47(9):1061--1067, 1994.

\bibitem{mioshi2006addenbrooke}
Eneida Mioshi, Kate Dawson, Joanna Mitchell, Robert Arnold, and John~R Hodges.
\newblock The addenbrooke's cognitive examination revised (ace-r): a brief
  cognitive test battery for dementia screening.
\newblock {\em International Journal of Geriatric Psychiatry: A journal of the
  psychiatry of late life and allied sciences}, 21(11):1078--1085, 2006.

\bibitem{mathuranath2000brief}
PS~Mathuranath, PJ~Nestor, GE~Berrios, Wojtek Rakowicz, and JR~Hodges.
\newblock A brief cognitive test battery to differentiate alzheimer's disease
  and frontotemporal dementia.
\newblock {\em Neurology}, 55(11):1613--1620, 2000.

\bibitem{hsieh2013validation}
Sharpley Hsieh, Samantha Schubert, Christopher Hoon, Eneida Mioshi, and John~R
  Hodges.
\newblock Validation of the addenbrooke's cognitive examination iii in
  frontotemporal dementia and alzheimer's disease.
\newblock {\em Dementia and geriatric cognitive disorders}, 36(3-4):242--250,
  2013.

\bibitem{alexopoulos2010validation}
Panagiotis Alexopoulos, A~Ebert, Tanja Richter-Schmidinger, E~Sch{\"o}ll,
  B~Natale, CA~Aguilar, P~Gourzis, M~Weih, R~Perneczky, J~Diehl-Schmid, et~al.
\newblock Validation of the german revised addenbrooke’s cognitive
  examination for detecting mild cognitive impairment, mild dementia in
  alzheimer’s disease and frontotemporal lobar degeneration.
\newblock {\em Dementia and geriatric cognitive disorders}, 29(5):448--456,
  2010.

\bibitem{fang2014validation}
Rong Fang, Gang Wang, Yue Huang, Jun-Peng Zhuang, Hui-Dong Tang, Ying Wang,
  Yu-Lei Deng, Wei Xu, Sheng-Di Chen, and Ru-Jing Ren.
\newblock Validation of the chinese version of addenbrooke's cognitive
  examination-revised for screening mild alzheimer's disease and mild cognitive
  impairment.
\newblock {\em Dementia and geriatric cognitive disorders}, 37(3-4):223--231,
  2014.

\bibitem{wong2013validation}
LL~Wong, CC~Chan, JL~Leung, CY~Yung, KK~Wu, SYY Cheung, and CLM Lam.
\newblock A validation study of the chinese-cantonese addenbrooke’s cognitive
  examination revised (c-acer).
\newblock {\em Neuropsychiatric disease and treatment}, 9:731, 2013.

\bibitem{dos2012validation}
Kelssy~Hitomi dos Santos~Kawata, Ryusaku Hashimoto, Yoshiyuki Nishio, Atsuko
  Hayashi, Nanayo Ogawa, Shigenori Kanno, Kotaro Hiraoka, Kayoko Yokoi, Osamu
  Iizuka, and Etsuro Mori.
\newblock A validation study of the japanese version of the addenbrooke’s
  cognitive examination-revised.
\newblock {\em Dementia and geriatric cognitive disorders extra}, 2(1):29--37,
  2012.

\bibitem{yoshida2012validation}
Hidenori Yoshida, Seishi Terada, Hajime Honda, Yuki Kishimoto, Naoya Takeda,
  Etsuko Oshima, Keisuke Hirayama, Osamu Yokota, and Yosuke Uchitomi.
\newblock Validation of the revised addenbrooke's cognitive examination (ace-r)
  for detecting mild cognitive impairment and dementia in a japanese
  population.
\newblock {\em International Psychogeriatrics}, 24(1):28--37, 2012.

\bibitem{kwak2010korean}
Yong~Tae Kwak, Youngsoon Yang, and Gyung~Whan Kim.
\newblock Korean addenbrooke's cognitive examination revised (k-acer) for
  differential diagnosis of alzheimer's disease and subcortical ischemic
  vascular dementia.
\newblock {\em Geriatrics \& gerontology international}, 10(4):295--301, 2010.

\bibitem{carvalho2010brazilian}
Viviane~Amaral Carvalho, Maira~Tonidandel Barbosa, and Paulo Caramelli.
\newblock Brazilian version of the addenbrooke cognitive examination-revised in
  the diagnosis of mild alzheimer disease.
\newblock {\em Cognitive and Behavioral Neurology}, 23(1):8--13, 2010.

\bibitem{garcia2006validation}
Alejandro Garc{\'\i}a-Caballero, I~Garc{\'\i}a-Lado, J~Gonz{\'a}lez-Hermida,
  MJ~Recimil, R~Area, F~Manes, S~Lamas, and GE~Berrios.
\newblock Validation of the spanish version of the addenbrooke's cognitive
  examination in a rural community in spain.
\newblock {\em International Journal of Geriatric Psychiatry: A journal of the
  psychiatry of late life and allied sciences}, 21(3):239--245, 2006.

\bibitem{gonccalves2015portuguese}
C{\'a}tia Gon{\c{c}}alves, Maria~Salom{\'e} Pinho, Vitor Cruz, Joana Pais,
  Helena Gens, F{\'a}tima Oliveira, Isabel Santana, Jos{\'e} Rente, and
  Jos{\'e}~Manuel Santos.
\newblock The portuguese version of addenbrooke’s cognitive
  examination--revised (ace-r) in the diagnosis of subcortical vascular
  dementia and alzheimer’s disease.
\newblock {\em Aging, Neuropsychology, and Cognition}, 22(4):473--485, 2015.

\bibitem{bastide2012addenbrooke}
Laure Bastide, Sandra De~Breucker, M~Van~den Berge, Patrick Fery, Thierry
  Pepersack, and Jean~Christophe Bier.
\newblock The addenbrooke’s cognitive examination revised is as effective as
  the original to detect dementia in a french-speaking population.
\newblock {\em Dementia and geriatric cognitive disorders}, 34(5-6):337--343,
  2012.

\bibitem{lee2008brief}
Jun-Young Lee, Dong~Woo Lee, Seong-Jin Cho, Duk~L Na, Hong~Jin Jeon, Shin-Kyum
  Kim, You~Ra Lee, Jung-Hae Youn, Miseon Kwon, Jae-Hong Lee, et~al.
\newblock Brief screening for mild cognitive impairment in elderly outpatient
  clinic: validation of the korean version of the montreal cognitive
  assessment.
\newblock {\em Journal of geriatric psychiatry and neurology}, 21(2):104--110,
  2008.

\bibitem{freitas2011montreal}
Sandra Freitas, M{\'a}rio~R Sim{\~o}es, Lara Alves, and Isabel Santana.
\newblock Montreal cognitive assessment (moca): normative study for the
  portuguese population.
\newblock {\em Journal of clinical and experimental neuropsychology},
  33(9):989--996, 2011.

\bibitem{luis2009cross}
Cheryl~A Luis, Andrew~P Keegan, and Michael Mullan.
\newblock Cross validation of the montreal cognitive assessment in community
  dwelling older adults residing in the southeastern us.
\newblock {\em International Journal of Geriatric Psychiatry: A journal of the
  psychiatry of late life and allied sciences}, 24(2):197--201, 2009.

\bibitem{memoria2013brief}
Cl{\'a}udia~M Mem{\'o}ria, M{\^o}nica~S Yassuda, Eduardo~Y Nakano, and
  Orestes~V Forlenza.
\newblock Brief screening for mild cognitive impairment: validation of the
  brazilian version of the montreal cognitive assessment.
\newblock {\em International Journal of Geriatric Psychiatry}, 28(1):34--40,
  2013.

\bibitem{lu2011montreal}
Jihui Lu, Dan Li, Fang Li, Aihong Zhou, Fen Wang, Xiumei Zuo, Xiang-Fei Jia,
  Haiqing Song, and Jianping Jia.
\newblock Montreal cognitive assessment in detecting cognitive impairment in
  chinese elderly individuals: a population-based study.
\newblock {\em Journal of geriatric psychiatry and neurology}, 24(4):184--190,
  2011.

\bibitem{yu2012beijing}
Jing Yu, Juan Li, and Xin Huang.
\newblock The beijing version of the montreal cognitive assessment as a brief
  screening tool for mild cognitive impairment: a community-based study.
\newblock {\em BMC psychiatry}, 12(1):156, 2012.

\bibitem{agrell1998clock}
Berit Agrell and Ove Dehlin.
\newblock The clock-drawing test.
\newblock {\em Age and ageing}, 27(3):399--403, 1998.

\bibitem{shulman2000clock}
Kenneth~I Shulman.
\newblock Clock-drawing: is it the ideal cognitive screening test?
\newblock {\em International journal of geriatric psychiatry}, 15(6):548--561,
  2000.

\bibitem{watson1993clock}
Yasmira~I Watson, Cynthia~L Arfken, and Stanley~J Birge.
\newblock Clock completion: an objective screening test for dementia.
\newblock {\em Journal of the American Geriatrics Society}, 41(11):1235--1240,
  1993.

\bibitem{danso2019application}
Samuel~O Danso, Graciela Muniz-Terrera, Saturnino Luz, Craig Ritchie, et~al.
\newblock Application of big data and artificial intelligence technologies to
  dementia prevention research: An opportunity for low-and-middle-income
  countries.
\newblock {\em Journal of Global Health}, 9(2), 2019.

\bibitem{williams2013machine}
Jennifer~A Williams, Alyssa Weakley, Diane~J Cook, and Maureen
  Schmitter-Edgecombe.
\newblock Machine learning techniques for diagnostic differentiation of mild
  cognitive impairment and dementia.
\newblock In {\em Workshops at the twenty-seventh AAAI conference on artificial
  intelligence}, 2013.

\bibitem{angelillo2019attentional}
Maria~Teresa Angelillo, Fabrizio Balducci, Donato Impedovo, Giuseppe Pirlo, and
  Gennaro Vessio.
\newblock Attentional pattern classification for automatic dementia detection.
\newblock {\em IEEE Access}, 7:57706--57716, 2019.

\bibitem{thabtah2020mobile}
Fadi Thabtah, Ella Mampusti, David Peebles, Raymund Herradura, et~al.
\newblock A mobile-based screening system for data analyses of early dementia
  traits detection.
\newblock {\em Journal of Medical Systems}, 44(1):24, 2020.

\bibitem{lim2012use}
Yen~Ying Lim, Kathryn~A Ellis, Karra Harrington, David Ames, Ralph~N Martins,
  Colin~L Masters, Christopher Rowe, Greg Savage, Cassandra Szoeke, David
  Darby, et~al.
\newblock Use of the cogstate brief battery in the assessment of alzheimer's
  disease related cognitive impairment in the australian imaging, biomarkers
  and lifestyle (aibl) study.
\newblock {\em Journal of clinical and experimental neuropsychology},
  34(4):345--358, 2012.

\bibitem{maruff2009validity}
Paul Maruff, Elizabeth Thomas, Lucette Cysique, Bruce Brew, Alex Collie, Peter
  Snyder, and Robert~H Pietrzak.
\newblock Validity of the cogstate brief battery: relationship to standardized
  tests and sensitivity to cognitive impairment in mild traumatic brain injury,
  schizophrenia, and aids dementia complex.
\newblock {\em Archives of Clinical Neuropsychology}, 24(2):165--178, 2009.

\bibitem{mielke2015performance}
Michelle~M Mielke, Mary~M Machulda, Clinton~E Hagen, Kelly~K Edwards, Rosebud~O
  Roberts, V~Shane Pankratz, David~S Knopman, Clifford~R Jack~Jr, and Ronald~C
  Petersen.
\newblock Performance of the cogstate computerized battery in the mayo clinic
  study on aging.
\newblock {\em Alzheimer's \& Dementia}, 11(11):1367--1376, 2015.

\bibitem{sahakian1992computerized}
Barbara~J Sahakian and AM~Owen.
\newblock Computerized assessment in neuropsychiatry using cantab: discussion
  paper.
\newblock {\em Journal of the Royal Society of Medicine}, 85(7):399, 1992.

\bibitem{barnett2015paired}
Jennifer~H Barnett, Andrew~D Blackwell, Barbara~J Sahakian, and Trevor~W
  Robbins.
\newblock The paired associates learning (pal) test: 30 years of cantab
  translational neuroscience from laboratory to bedside in dementia research.
\newblock {\em Translational neuropsychopharmacology}, pages 449--474, 2015.

\bibitem{yu2015development}
Ke~Yu, Shangang Zhang, Qingsong Wang, Xiaofei Wang, Yang Qin, Jian Wang,
  Congyang Li, Yuxian Wu, Weiwen Wang, and Hang Lin.
\newblock Development of a computerized tool for the chinese version of the
  montreal cognitive assessment for screening mild cognitive impairment.
\newblock {\em International psychogeriatrics}, 27(2):213, 2015.

\bibitem{memoria2014contributions}
Cl{\'a}udia~M Mem{\'o}ria, M{\^o}nica~S Yassuda, Eduardo~Y Nakano, and
  Orestes~V Forlenza.
\newblock Contributions of the computer-administered neuropsychological screen
  for mild cognitive impairment (cans-mci) for the diagnosis of mci in brazil.
\newblock {\em International psychogeriatrics}, 26(9):1483, 2014.

\bibitem{shigemori2016dementia}
Tomoaki Shigemori, Hiroharu Kawanaka, Yulia Hicks, Rossi Setchi, Haruhiko
  Takase, and Shinji Tsuruoka.
\newblock Dementia detection using weighted direction index histograms and svm
  for clock drawing test.
\newblock {\em Procedia Computer Science}, 96:1240--1248, 2016.

\bibitem{bennasar2014cascade}
Mohamed Bennasar, Rossitza Setchi, Yulia Hicks, and Antony Bayer.
\newblock Cascade classification for diagnosing dementia.
\newblock In {\em 2014 IEEE International Conference on Systems, Man, and
  Cybernetics (SMC)}, pages 2535--2540. IEEE, 2014.

\bibitem{impedovo2019handwriting}
Donato Impedovo, Giuseppe Pirlo, Gennaro Vessio, and Maria~Teresa Angelillo.
\newblock A handwriting-based protocol for assessing neurodegenerative
  dementia.
\newblock {\em Cognitive Computation}, 11(4):576--586, 2019.

\bibitem{lerche2016prospective}
Stefanie Lerche, Kathrin Brockmann, Andrea Pilotto, Isabel Wurster, Ulrike
  S{\"u}nkel, Markus~A Hobert, Anna-Katharina von Thaler, Claudia Schulte, Erik
  Stoops, Hugo Vanderstichele, et~al.
\newblock Prospective longitudinal course of cognition in older subjects with
  mild parkinsonian signs.
\newblock {\em Alzheimer's research \& therapy}, 8(1):42, 2016.

\bibitem{buracchio2010trajectory}
Teresa Buracchio, Hiroko~H Dodge, Diane Howieson, Dara Wasserman, and Jeffrey
  Kaye.
\newblock The trajectory of gait speed preceding mild cognitive impairment.
\newblock {\em Archives of neurology}, 67(8):980--986, 2010.

\bibitem{aggarwal2006motor}
Neelum~T Aggarwal, Robert~S Wilson, Todd~L Beck, Julia~L Bienias, and David~A
  Bennett.
\newblock Motor dysfunction in mild cognitive impairment and the risk of
  incident alzheimer disease.
\newblock {\em Archives of Neurology}, 63(12):1763--1769, 2006.

\bibitem{boyle2005parkinsonian}
PA~Boyle, RS~Wilson, NT~Aggarwal, Z~Arvanitakis, J~Kelly, JL~Bienias, and
  DA~Bennett.
\newblock Parkinsonian signs in subjects with mild cognitive impairment.
\newblock {\em Neurology}, 65(12):1901--1906, 2005.

\bibitem{wilson2003parkinsonianlike}
Robert~S Wilson, Julie~A Schneider, Julia~L Bienias, Denis~A Evans, and David~A
  Bennett.
\newblock Parkinsonianlike signs and risk of incident alzheimer disease in
  older persons.
\newblock {\em Archives of neurology}, 60(4):539--544, 2003.

\bibitem{camicioli1998motor}
Richard Camicioli, Diane Howieson, Barry Oken, G~Sexton, and Jeffrey Kaye.
\newblock Motor slowing precedes cognitive impairment in the oldest old.
\newblock {\em Neurology}, 50(5):1496--1498, 1998.

\bibitem{beauchet2016association}
Olivier Beauchet, Gilles Allali, C{\'e}dric Annweiler, and Joe Verghese.
\newblock Association of motoric cognitive risk syndrome with brain volumes:
  results from the gait study.
\newblock {\em Journals of Gerontology Series A: Biomedical Sciences and
  Medical Sciences}, 71(8):1081--1088, 2016.

\bibitem{buckley2019role}
Christopher Buckley, Lisa Alcock, R{\'\i}ona McArdle, Rana Zia~Ur Rehman,
  Silvia Del~Din, Claudia Mazz{\`a}, Alison~J Yarnall, and Lynn Rochester.
\newblock The role of movement analysis in diagnosing and monitoring
  neurodegenerative conditions: Insights from gait and postural control.
\newblock {\em Brain sciences}, 9(2):34, 2019.

\bibitem{mollica2019early}
Maria~A Mollica, Adri{\`a} Tort-Merino, Jordi Navarra, Irune
  Fern{\'a}ndez-Prieto, Natalia Valech, Jaume Olives, Mar{\'\i}a Le{\'o}n,
  Alberto Lle{\'o}, Pablo Mart{\'\i}nez-Lage, Raquel S{\'a}nchez-Valle, et~al.
\newblock Early detection of subtle motor dysfunction in cognitively normal
  subjects with amyloid-$\beta$ positivity.
\newblock {\em Cortex}, 121:117--124, 2019.

\bibitem{mc2020differentiating}
R{\'\i}ona Mc~Ardle, Silvia Del~Din, Brook Galna, Alan Thomas, and Lynn
  Rochester.
\newblock Differentiating dementia disease subtypes with gait analysis:
  feasibility of wearable sensors?
\newblock {\em Gait \& posture}, 76:372--376, 2020.

\bibitem{chung2012gait}
Pau-Choo Chung, Yu-Liang Hsu, Chun-Yao Wang, Chien-Wen Lin, Jeen-Shing Wang,
  and Ming-Chyi Pai.
\newblock Gait analysis for patients with alzheimer's disease using a triaxial
  accelerometer.
\newblock In {\em 2012 IEEE International Symposium on Circuits and Systems
  (ISCAS)}, pages 1323--1326. IEEE, 2012.

\bibitem{annweiler29motoric}
Cedric Annweiler, Emmeline Ayers, Nir Barzilai, Olivier Beauchet, David~A
  Bennett, Stephanie~A Bridenbaugh, Aron~S Buchman, Michele~L Callisaya,
  Richard Camicioli, Benjamin Capistrant, et~al.
\newblock Motoric cognitive risk syndrome: Multicountry prevalence and dementia
  risk.

\bibitem{sano2019detection}
Yuko Sano, Ying Yin, Tomohiko Mizuguchi, and Akihiko Kandori.
\newblock Detection of abnormal segments in finger tapping waveform using
  one-class svm.
\newblock In {\em 2019 41st Annual International Conference of the IEEE
  Engineering in Medicine and Biology Society (EMBC)}, pages 1378--1381. IEEE,
  2019.

\bibitem{liang2019real}
Xing Liang, Epaminondas Kapetanios, Bencie Woll, and Anastassia Angelopoulou.
\newblock Real time hand movement trajectory tracking for enhancing dementia
  screening in ageing deaf signers of british sign language.
\newblock In {\em International Cross-Domain Conference for Machine Learning
  and Knowledge Extraction}, pages 377--394. Springer, 2019.

\bibitem{yu2019characterization}
Nan-Ying Yu and Shao-Hsia Chang.
\newblock Characterization of the fine motor problems in patients with
  cognitive dysfunction--a computerized handwriting analysis.
\newblock {\em Human movement science}, 65:71--79, 2019.

\bibitem{crutcher2009eye}
Michael~D Crutcher, Rose Calhoun-Haney, Cecelia~M Manzanares, James~J Lah,
  Allan~I Levey, and Stuart~M Zola.
\newblock Eye tracking during a visual paired comparison task as a predictor of
  early dementia.
\newblock {\em American Journal of Alzheimer's Disease \& Other
  Dementias{\textregistered}}, 24(3):258--266, 2009.

\bibitem{anderson2013eye}
Tim~J Anderson and Michael~R MacAskill.
\newblock Eye movements in patients with neurodegenerative disorders.
\newblock {\em Nature Reviews Neurology}, 9(2):74--85, 2013.

\bibitem{chatterjee2018eye}
Debatri Chatterjee, Rahul~Dasharath Gavas, Kingshuk Chakravarty, Aniruddha
  Sinha, and Uttama Lahiri.
\newblock Eye movements-an early marker of cognitive dysfunctions.
\newblock In {\em 2018 40th Annual International Conference of the IEEE
  Engineering in Medicine and Biology Society (EMBC)}, pages 4012--4016. IEEE,
  2018.

\bibitem{currie1988eye}
Jon Currie, Elizabeth Benson, Ben Ramsden, Michael Perdices, and David Cooper.
\newblock Eye movement abnormalities as a predictor of the acquired
  immunodeficiency syndrome dementia complex.
\newblock {\em Archives of neurology}, 45(9):949--953, 1988.

\bibitem{beltran2018computational}
Jessica Beltr{\'a}n, Mireya~S Garc{\'\i}a-V{\'a}zquez, Jenny Benois-Pineau,
  Luis~Miguel Gutierrez-Robledo, and Jean-Fran{\c{c}}ois Dartigues.
\newblock Computational techniques for eye movements analysis towards
  supporting early diagnosis of alzheimer’s disease: a review.
\newblock {\em Computational and mathematical methods in medicine}, 2018, 2018.

\bibitem{fraser2017analysis}
Kathleen~C Fraser, Kristina~Lundholm Fors, Dimitrios Kokkinakis, and Arto
  Nordlund.
\newblock An analysis of eye-movements during reading for the detection of mild
  cognitive impairment.
\newblock In {\em Proceedings of the 2017 Conference on Empirical Methods in
  Natural Language Processing}, pages 1016--1026, 2017.

\bibitem{zhang2016monitoring}
Yanxia Zhang, Thomas Wilcockson, Kwang~In Kim, Trevor Crawford, Hans Gellersen,
  and Pete Sawyer.
\newblock Monitoring dementia with automatic eye movements analysis.
\newblock In {\em Intelligent Decision Technologies 2016}, pages 299--309.
  Springer, 2016.

\bibitem{hopper2007service}
Tammy Hopper, Stuart Cleary, Bruce Oddson, Mary~Jo Donnelly, and Shawna Elgar.
\newblock Service delivery for older canadians with dementia: A survey of
  speech-language pathologists.
\newblock {\em Revue canadienne d’orthophonie et d’audiologie-Vol},
  31(3):115, 2007.

\bibitem{kempler2008language}
Daniel Kempler and Mira Goral.
\newblock Language and dementia: Neuropsychological aspects.
\newblock {\em Annual review of applied linguistics}, 28:73, 2008.

\bibitem{haider2019assessment}
Fasih Haider, Sofia De~La~Fuente, and Saturnino Luz.
\newblock An assessment of paralinguistic acoustic features for detection of
  alzheimer's dementia in spontaneous speech.
\newblock {\em IEEE Journal of Selected Topics in Signal Processing},
  14(2):272--281, 2019.

\bibitem{chien2019automatic}
Yi-Wei Chien, Sheng-Yi Hong, Wen-Ting Cheah, Li-Hung Yao, Yu-Ling Chang, and
  Li-Chen Fu.
\newblock An automatic assessment system for alzheimer’s disease based on
  speech using feature sequence generator and recurrent neural network.
\newblock {\em Scientific Reports}, 9(1):1--10, 2019.

\bibitem{liu2019dementia}
Zhaoci Liu, Zhiqiang Guo, Zhenhua Ling, Shijin Wang, Lingjing Jin, and Yunxia
  Li.
\newblock Dementia detection by analyzing spontaneous mandarin speech.
\newblock In {\em 2019 Asia-Pacific Signal and Information Processing
  Association Annual Summit and Conference (APSIPA ASC)}, pages 289--296. IEEE,
  2019.

\bibitem{pan2019automatic}
Yilin Pan, Bahman Mirheidari, Markus Reuber, Annalena Venneri, Daniel
  Blackburn, and Heidi Christensen.
\newblock Automatic hierarchical attention neural network for detecting ad.
\newblock In {\em Interspeech}, pages 4105--4109, 2019.

\bibitem{tanaka2017detecting}
Hiroki Tanaka, Hiroyoshi Adachi, Norimichi Ukita, Manabu Ikeda, Hiroaki Kazui,
  Takashi Kudo, and Satoshi Nakamura.
\newblock Detecting dementia through interactive computer avatars.
\newblock {\em IEEE journal of translational engineering in health and
  medicine}, 5:1--11, 2017.

\bibitem{mirheidari2018detecting}
Bahman Mirheidari, Daniel Blackburn, Traci Walker, Annalena Venneri, Markus
  Reuber, and Heidi Christensen.
\newblock Detecting signs of dementia using word vector representations.
\newblock In {\em Interspeech}, pages 1893--1897, 2018.

\bibitem{luz2018method}
Saturnino Luz, Sofia de~la Fuente, and Pierre Albert.
\newblock A method for analysis of patient speech in dialogue for dementia
  detection.
\newblock {\em arXiv preprint arXiv:1811.09919}, 2018.

\bibitem{ujiro2018detection}
Tsuyoki Ujiro, Hiroki Tanaka, Hiroyoshi Adachi, Hiroaki Kazui, Manabu Ikeda,
  Takashi Kudo, and Satoshi Nakamura.
\newblock Detection of dementia from responses to atypical questions asked by
  embodied conversational agents.
\newblock In {\em Interspeech}, pages 1691--1695, 2018.

\bibitem{chien2018assessment}
Yi-Wei Chien, Sheng-Yi Hong, Wen-Ting Cheah, Li-Chen Fu, and Yu-Ling Chang.
\newblock An assessment system for alzheimer's disease based on speech using a
  novel feature sequence design and recurrent neural network.
\newblock In {\em 2018 IEEE International Conference on Systems, Man, and
  Cybernetics (SMC)}, pages 3289--3294. IEEE, 2018.

\bibitem{weiner2018investigating}
Jochen Weiner, Miguel Angrick, Srinivasan Umesh, and Tanja Schultz.
\newblock Investigating the effect of audio duration on dementia detection
  using acoustic features.
\newblock In {\em Interspeech}, pages 2324--2328, 2018.

\bibitem{mirheidari2019dementia}
Bahman Mirheidari, Daniel Blackburn, Traci Walker, Markus Reuber, and Heidi
  Christensen.
\newblock Dementia detection using automatic analysis of conversations.
\newblock {\em Computer Speech \& Language}, 53:65--79, 2019.

\bibitem{weiner2017manual}
Jochen Weiner, Mathis Engelbart, and Tanja Schultz.
\newblock Manual and automatic transcriptions in dementia detection from
  speech.
\newblock In {\em INTERSPEECH}, pages 3117--3121, 2017.

\bibitem{bullard2016towards}
Joseph Bullard, Cecilia~Ovesdotter Alm, Xumin Liu, Qi~Yu, and Ruben~A Proano.
\newblock Towards early dementia detection: fusing linguistic and
  non-linguistic clinical data.
\newblock In {\em Proceedings of the Third Workshop on Computational
  Linguistics and Clinical Psychology}, pages 12--22, 2016.

\bibitem{shao2019detection}
Yijun Shao, Qing~T Zeng, Kathryn~K Chen, Andrew Shutes-David, Stephen~M
  Thielke, and Debby~W Tsuang.
\newblock Detection of probable dementia cases in undiagnosed patients using
  structured and unstructured electronic health records.
\newblock {\em BMC medical informatics and decision making}, 19(1):128, 2019.

\bibitem{balbim2020evaluation}
Guilherme~M Balbim, Ashley~M Maldonado, Amy Early, Lesley Steinman, Kristin
  Harkins, and David~X Marquez.
\newblock Evaluation of public health messages promoting early detection of
  dementia among adult latinos with a living older adult parental figure.
\newblock {\em Hispanic Health Care International}, page 1540415320908535,
  2020.

\bibitem{pellegrini2018machine}
Enrico Pellegrini, Lucia Ballerini, Maria del C~Valdes Hernandez, Francesca~M
  Chappell, Victor Gonz{\'a}lez-Castro, Devasuda Anblagan, Samuel Danso, Susana
  Mu{\~n}oz-Maniega, Dominic Job, Cyril Pernet, et~al.
\newblock Machine learning of neuroimaging for assisted diagnosis of cognitive
  impairment and dementia: A systematic review.
\newblock {\em Alzheimer's \& Dementia: Diagnosis, Assessment \& Disease
  Monitoring}, 10:519--535, 2018.

\bibitem{mathotaarachchi2017identifying}
Sulantha Mathotaarachchi, Tharick~A Pascoal, Monica Shin, Andrea~L Benedet,
  Min~Su Kang, Thomas Beaudry, Vladimir~S Fonov, Serge Gauthier, Pedro
  Rosa-Neto, Alzheimer's Disease~Neuroimaging Initiative, et~al.
\newblock Identifying incipient dementia individuals using machine learning and
  amyloid imaging.
\newblock {\em Neurobiology of aging}, 59:80--90, 2017.

\bibitem{mueller2005alzheimer}
Susanne~G Mueller, Michael~W Weiner, Leon~J Thal, Ronald~C Petersen, Clifford
  Jack, William Jagust, John~Q Trojanowski, Arthur~W Toga, and Laurel Beckett.
\newblock The alzheimer's disease neuroimaging initiative.
\newblock {\em Neuroimaging Clinics}, 15(4):869--877, 2005.

\bibitem{marcus2007open}
Daniel~S Marcus, Tracy~H Wang, Jamie Parker, John~G Csernansky, John~C Morris,
  and Randy~L Buckner.
\newblock Open access series of imaging studies (oasis): cross-sectional mri
  data in young, middle aged, nondemented, and demented older adults.
\newblock {\em Journal of cognitive neuroscience}, 19(9):1498--1507, 2007.

\bibitem{bidani2019dementia}
Amen Bidani, Mohamed~Salah Gouider, and Carlos~M Travieso-Gonz{\'a}lez.
\newblock Dementia detection and classification from mri images using deep
  neural networks and transfer learning.
\newblock In {\em International Work-Conference on Artificial Neural Networks},
  pages 925--933. Springer, 2019.

\bibitem{iizuka2019deep}
Tomomichi Iizuka, Makoto Fukasawa, and Masashi Kameyama.
\newblock Deep-learning-based imaging-classification identified cingulate
  island sign in dementia with lewy bodies.
\newblock {\em Scientific reports}, 9(1):1--9, 2019.

\bibitem{islam2018early}
Jyoti Islam and Yanqing Zhang.
\newblock Early diagnosis of alzheimer's disease: A neuroimaging study with
  deep learning architectures.
\newblock In {\em Proceedings of the IEEE conference on computer vision and
  pattern recognition workshops}, pages 1881--1883, 2018.

\bibitem{yiugit2020applying}
Altu{\u{g}} Y{\.I}{\u{G}}{\.I}T and Zerrin I{\c{S}}IK.
\newblock Applying deep learning models to structural mri for stage prediction
  of alzheimer’s disease.
\newblock {\em Turkish Journal of Electrical Engineering \& Computer Sciences},
  28:196--210, 2020.

\bibitem{backstrom2018efficient}
Karl B{\"a}ckstr{\"o}m, Mahmood Nazari, Irene Yu-Hua Gu, and Asgeir~Store
  Jakola.
\newblock An efficient 3d deep convolutional network for alzheimer's disease
  diagnosis using mr images.
\newblock In {\em 2018 IEEE 15th International Symposium on Biomedical Imaging
  (ISBI 2018)}, pages 149--153. IEEE, 2018.

\bibitem{kompanek2019volumetrie}
Matej Kompanek, Martin Tamajka, and Wanda Benesova.
\newblock Volumetrie data augmentation as an effective tool in mri
  classification using 3d convolutional neural network.
\newblock In {\em 2019 International Conference on Systems, Signals and Image
  Processing (IWSSIP)}, pages 115--119. IEEE, 2019.

\bibitem{lotfi2012smart}
Ahmad Lotfi, Caroline Langensiepen, Sawsan~M Mahmoud, and Mohammad~Javad
  Akhlaghinia.
\newblock Smart homes for the elderly dementia sufferers: identification and
  prediction of abnormal behaviour.
\newblock {\em Journal of ambient intelligence and humanized computing},
  3(3):205--218, 2012.

\bibitem{batista2016wandering}
Edgar Batista, Fran Casino, and Agusti Solanas.
\newblock On wandering detection methods in context-aware scenarios.
\newblock In {\em 2016 7th International Conference on Information,
  Intelligence, Systems \& Applications (IISA)}, pages 1--6. IEEE, 2016.

\bibitem{khodabandehloo2020collaborative}
Elham Khodabandehloo and Daniele Riboni.
\newblock Collaborative trajectory mining in smart-homes to support early
  diagnosis of cognitive decline.
\newblock {\em IEEE Transactions on Emerging Topics in Computing}, 2020.

\bibitem{paolini2017human}
Giacomo Paolini, Diego Masotti, Alessandra Costanzo, Elena Borelli, Lorenzo
  Chiari, Silvia Imbesi, Michele Marchi, and Giuseppe Mincolelli.
\newblock Human-centered design of a smart “wireless sensor network
  environment” enhanced with movement analysis system and indoor positioning
  qualifications.
\newblock In {\em 2017 IEEE MTT-S International Microwave Workshop Series on
  Advanced Materials and Processes for RF and THz Applications (IMWS-AMP)},
  pages 1--3. IEEE, 2017.

\bibitem{stavropoulos2016semantic}
Thanos~G Stavropoulos, Georgios Meditskos, Stelios Andreadis, Konstantinos
  Avgerinakis, Katerina Adam, and Ioannis Kompatsiaris.
\newblock Semantic event fusion of computer vision and ambient sensor data for
  activity recognition to support dementia care.
\newblock {\em Journal of Ambient Intelligence and Humanized Computing}, pages
  1--16, 2016.

\bibitem{poole2017motor}
Matthew~L Poole, Amy Brodtmann, David Darby, and Adam~P Vogel.
\newblock Motor speech phenotypes of frontotemporal dementia, primary
  progressive aphasia, and progressive apraxia of speech.
\newblock {\em Journal of Speech, Language, and Hearing Research},
  60(4):897--911, 2017.

\end{thebibliography}

\end{document}